%% file: ArxivVersion2.tex
\documentclass{article}
\usepackage{arxiv}
\usepackage[utf8]{inputenc} % allow utf-8 input
\usepackage[T1]{fontenc}    % use 8-bit T1 fonts
\usepackage{hyperref}       % hyperlinks
\usepackage{url}            % simple URL typesetting
\usepackage{booktabs}       % professional-quality tables
\usepackage{amsfonts}       % blackboard math symbols
\usepackage{nicefrac}       % compact symbols for 1/2, etc.
\usepackage{microtype}      % microtypography
\usepackage{lipsum}		% Can be removed after putting your text content
\usepackage{graphicx}
\usepackage{amsmath,amssymb,amsfonts}
\usepackage{algorithmic}
\usepackage{graphicx}
\usepackage{textcomp}
\usepackage{xcolor}
\usepackage{booktabs} % For formal tables
\usepackage{graphicx}
\usepackage{epstopdf}

\usepackage{url}
\usepackage{algorithmic}
\usepackage{hyperref}
\usepackage{multirow}
\usepackage{caption}
\usepackage{amssymb}
\usepackage{amsfonts}
\usepackage{float}
\usepackage{bm}
\usepackage{dsfont}
\usepackage[T1]{fontenc}
\usepackage[left,modulo]{lineno}
\usepackage[ruled, vlined, linesnumbered, longend]{algorithm2e}
\usepackage{psfrag}
\usepackage{subfigure}
\usepackage{fancyhdr}
\usepackage{eucal}
\usepackage[english]{babel}
\usepackage{ifpdf}
\usepackage{setspace}
\usepackage{textcomp}
\usepackage{tikz}
\usepackage{tikz-inet}
\usetikzlibrary{calc,shapes.geometric}
\usetikzlibrary{shapes,snakes,arrows,automata} 
\usetikzlibrary{patterns}

\title{Re-visiting Reservoir Computing architectures optimized by Evolutionary Algorithms
\thanks{This work was supported by the GACR-Czech Science Foundation project No. 21-33574K ``Lifelong Machine Learning on Data Streams''. Accepted manuscript to the 14th World Congress on Nature and Biologically Inspired Computing (NaBIC), Seattle, WA, United States, December 14-16, 2022. A revised manuscript will be published in the conference proceedings by Springer in the Lecture Notes in Networks and Systems.
}
}

\author{\href{https://orcid.org/0000-0002-9172-0155}{\includegraphics[scale=0.06]{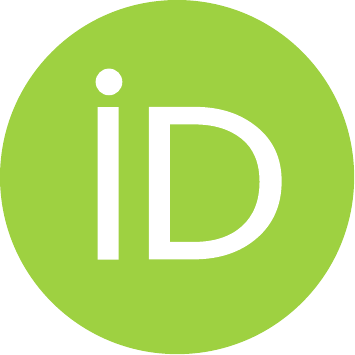}\hspace{1mm}Sebasti\'an Basterrech}\\
	Faculty of Electrical Engineering and Computer Science\\
	V\v{S}B-Technical University of Ostrava\\
	Ostrava, Czech Republic\\
	\texttt{Sebastian.Basterrech@vsb.cz}
	\And
	\href{https://orcid.org/0000-0002-9043-8641}{\includegraphics[scale=0.06]{orcid.pdf}\hspace{1mm}Tarun Kumar Sharma} \\
	Department of Computer Science and Engineering\\ 
	Shobhit University\\
	Saharanpur, India\\
	\texttt{taruniitr1@gmail.com}\\
}

\date{}
\renewcommand{\headeright}{$\;$}
\renewcommand{\undertitle}{(Accepted manuscript at NABIC'2022)}
\renewcommand{\shorttitle}{Re-visiting RC architectures optimized by EAs}

\begin{document}

\include{macrosBaster}
\newcommand{\op}{w^+}
\newcommand{\om}{w^-}
\newcommand{\Rho}{\mathrm{P}}
\newcommand{\wresp}{\mathbf{w}^+}
\newcommand{\wresn}{\mathbf{w}^-}
\renewcommand{\wre}{\mathbf{W}^{\rm{r}}}
\renewcommand{\wfb}{\mathbf{W}^{\rm{fb}}}
\newcommand{\ve}{\mathbf{v}}
\newcommand{\noise}{\bm{\varepsilon}}
\newcommand{\idct}{\phi}
\newcommand{\chr}{\kappa}
\newcommand{\phenotype}{\chi}
\newcommand{\mapping}{\Psi}
\renewcommand{\dim}{M}
\renewcommand{\Na}{K}
\newcommand{\Nx}{N}
\newcommand{\Ny}{L}
\newcommand{\State}{\mathbf{S}}
\newcommand{\pos}{\mathbf{p}}
\renewcommand{\vec}{\mathbf{v}}
\newcommand{\best}{\mathbf{b}}

% Used for displaying a sample figure. If possible, figure files should
% If you use the hyperref package, please uncomment the following line
% to display URLs in blue roman font according to Springer's eBook style:
% \renewcommand\UrlFont{\color{blue}\rmfamily}

% --- myCorrs
%     to make modifications visible
%     needs \usepackage{color} or \usepackage{xcolor}
%
\newif\ifnotes\notestrue
\def\boxnote#1#2{\ifnotes\fbox{\footnote{\ }}\ \footnotetext{ From #1: #2}\fi}

\def\bgr#1{\boxnote{Gerardo}{\color{red}#1}}
\def\hgr#1{}
\newcommand{\mgr}[1]{{\color{red}#1}}

\def\bsb#1{\boxnote{Sebasti\'an}{\bf\color{blue}#1}}
\renewcommand{\msb}[1]{{\bf\color{blue}#1}}

\newcommand{\todo}[1]{{\color{brown}\textsf{\bf [#1]}}}

\newcommand{\forceindent}{\leavevmode{\parindent=2em\indent}}

\newcommand{\ti}{\!\times\!}
\newcommand{\dimwa}{\Nx\ti\Na}
\newcommand{\dimwr}{\Nx\ti\Nx}
\newcommand{\dimwfb}{\Nx\ti\Ny}
\newcommand{\dimwo}{\Nx\ti(\Na\!+\!\Ny)}

\newcommand{\espacio}[1]{}

\newcommand{\vu}{\mathbf{u}}
\renewcommand{\vy}{\mathbf{y}}
\newcommand{\vz}{\mathbf{z}}
\newcommand{\U}{\mathbb{U}}
\renewcommand{\Y}{\mathbb{Y}}
\newcommand{\M}{\mathcal{M}}
\newcommand{\pred}{\hat{\mathbf{y}}}
\newcommand{\param}{\bm{\theta}}

\maketitle

\begin{abstract}
For many years, Evolutionary Algorithms (EAs) have been applied to improve Neural Networks (NNs) architectures.
They have been used for solving different problems, such as training the networks (adjusting the weights), designing network topology, optimizing global parameters, and selecting features.
Here, we provide a systematic brief survey about applications of the EAs on the specific domain of the recurrent NNs named Reservoir Computing (RC).
At the beginning of the 2000s, the RC paradigm appeared as a good option for employing recurrent NNs without dealing with the inconveniences of the training algorithms.
RC models use a nonlinear dynamic system, with fixed recurrent neural network named the \textit{reservoir}, and learning process is restricted to adjusting a linear parametric function. %so the performance of learning is fast and precise.
However, an RC model has several hyper-parameters, therefore EAs are helpful tools to figure out optimal RC architectures.
We provide an overview of the results on the area, discuss novel advances, and we present our vision regarding the new trends and still open questions.
\end{abstract}

% keywords can be removed

\newcommand{\Space}{\mathbb{S}}
\section{Introduction} 
\label{Introduction}
Recurrent Neural Network (RNNs) are ``powerful'' tools for modeling time series and solving machine learning problems with sequential data. 
A main characteristic of neural circuits is their highly recurrent structure that is composed by different types of neurons and synapses~\cite{Maass2016Brain}. Then, even though an RNN is a rough simplification of the neural system, the recurrent nets preserve biological plausibility~\cite{JaegerConceptors2018,Jaeger09}. 
They have also other advantages with respect of feed-forward models, the most mentioned in the literature are~\cite{Schmidhuber15,Jaeger09}: ``good'' theoretical properties, ``relative'' success in modelling chaotic systems, and real-world applications. 
However, RNNs are still not so popular like other families of Neural Networks (NNs).
Probably, one of the reason for this, it is the fact that it is still hard to efficiently train large recurrent networks.
Even though, NN optimization is one of the most investigated problems in the Machine Learning (ML) area, the problem has still open questions and it is still drawing a lot of interest. 
Specially, in the case of optimizing large recurrent topologies due to the complexity and difficulties presented in this problem.
%
%The difficulties presented in the learning of the weights are often referred under the name of  ``vanishing-exploiding'' gradient problem~\cite{Pascanu13}. 
%
In summary, the main problems in the training algorithms for RNNs are: not guaranteed convergence, long training times, and it is very difficult to store information in internal states for long periods (long-range memory)~\cite{Jaeger09,Graves12}.

At the beginning of 2000s has started a new approach for designing and training RNNs. 
The original motivation was to take advantage of the benefits of the recurrences for memorizing the sequential data, avoiding the problems of training recurrent networks. 
This approach was simultaneously developed with names of  \textit{Liquid State Machine (LSM)} (LSM)~\cite{Maass99} and \textit{Echo State Network} (ESN)~\cite{Jaeger01}. 
Both models and some of their variations converged in a new trend named \textit{Reservoir Computing} (RC)~\cite{Verstraeten07}.
An RC model is a NN with a recurrent hidden-to-hidden structure (so-called \textit{reservoir}) composed by fixed hidden-hidden weights, and the rest of the model weights (\textit{readout weights}) are trained using classic supervised learning.
The two canonical models (ESN and LSM) compute the readout weights using a simple linear regression.
It is a distinctive characteristic of the method, that the weights of the recurrent topology are randomly initialized and kept fixed during the learning process.
The training algorithm only focuses on adjusting a selected group of weights in order of having a fast and reliable  optimization step.
The RC models share this principle with other NN families, e.g. Extreme Learning Machines (ELM)~\cite{Huang2006},  where the hidden weights of first layers need not be tuned.
Both approaches are based on a static non-linear random projection from the input space to a feature space, and the training phase is robust and fast. 
Even though the apparent simplicity of the RC models, they have obtained outstanding performances in several real applications~\cite{Verstraeten07,Jaeger09,JaegerConceptors2018,GALLICCHIO201787,Li2012,BasterrechNCAA2019}.
Specially, the large popularity and recognition have started after the record-breaking success in modelling the Mackey-Glass attractor~\cite{JaegerScience04}.
%Therefore, a good property of the RC model is that the learning algorithm is very fast and robust.
%

The initial NN design always has impact on the performance of the model. In the case of an RC model, this effect may be even more dramatic. Because the reservoir weights are frozen during the training process.
The most common approach of defining the recurrent weights is using a random distribution, and then to scale them in order of satisfying some algebraic restrictions. 
In addition to the scaling of the recurrent weights, the model has other initial operations and global parameters.
The most relevant global parameters are related to the dimensionality of the feature space, stability of the recurrent  dynamics, memorization capacity of the system, non-linear transformations of the activation functions, and type of the readout mapping~\cite{JaegerScience04,BasterrechIJCNN2017,Jaeger09,Butcher2013}.
Most research effort has gone for understanding the impact of those parameters. For more information see~\cite{Hart2020,Jaeger09,Tanaka2019,Wainrib15}.
%{Jaeger2001,Manjunath2013,Yildiz2012,Wainrib15,Hart2020}. 
%
However, to define good hyper-parameters of an RC model is still often made by experience using a grid search strategy. Even, sometimes it is made partially by brute force. 

Around ten years ago, several works have applied the power of Evolutionary Algorithms (EAs) for improving the RC performance. 
The algorithms have been used for finding the global RC parameters mentioned above. Another work direction has been to developed hybrid approaches that combine the evolutionary aspects and the RC characteristics.
In this article, we provide a comprehensive global overview of the RC literature that contains EA. We analyze how EAs  are used, we present a critical vision of the area and discuss the new trends. The survey is not exhaustive, but  we cover many of the most important works on the area.
There are several surveys about RC models, for instance~\cite{Tanaka2019, Jaeger09, Sun2020}. However, so far as we know, this is the first overview regarding the specific sub-field of RC models where EAs area applied for optimizing the model architecture.

The remainder of this article is organized as follows. In the next section, we formalize the context where in the RC models are applied. Thereafter, we present the mathematical specification of the Echo State Network, its properties and limitations.
This is followed by the presentation of recent developed RC architectures.
Section~\ref{EA} describes the applications of EAs for improving the RC models.
We ends with a final discussion about the still open questions on the area.

\section{Reservoir Computing Methods}
\subsection{Context}
Let consider a learning dataset in a supervised context composed by $N$ input-output pairs $(\va,\vy)$, where the input space has dimension ${\Na}$ and output space has dimension ${\Ny}$.
For simplicity, we consider indexed instances in discrete spaces.
RC models have been used for solving both temporal and non-temporal tasks~\cite{Jaeger09}.
By temporal tasks or temporal learning, we refer to the situation when any instance $\va(t)$ depends on the data $\va(t+\Delta)$ for some non-large $\Delta$ value. 
It is the common situation of time-series, or sequential data.
By non-temporal tasks, we refer by the problems where the data instances are independent of each other.
In temporal learning, the common goal is to predict the next value of the sequence (or input signal) given the previous values, i.e. to learn $\vy(t)$ having information of $\va(t),\va(t-1),\va(t-1), \ldots$.
In non-temporal learning, the task is to learn $\y(t)$ given the current input $\va(t)$.
Here, we focus mainly on the case of temporal learning that is much more suitable for RNNs.
However, RC models also have been used on time-independent data~\cite{Alex09}. Furthermore, it has also been used in  time-series classification where  each input instance (a signal in this case) is independent of each other~\cite{Qianli2016}.
\subsection{Recurrent Neural Networks}

Given an input signal $\va(t)\in \R^{\Na}$ the simple case of RNN is described by a $N$-dimensional state $x(t)\in\R^{\Nx}$ following the recurrence:
\begin{equation}
    \label{recurrentState}
    \x(t)=f(\x(t-1), \va(t), \bm{\theta_1}), 
\end{equation}
where $f(\cdot)$ is a regular parametric function (e.g. a linear combination following by a regular function such as hyperbolic tangent, sigmoid forms, etc.), $\bm{\theta}$ is the weight collection (model parameters).
Note that, expression~(\ref{recurrentState}) at each time step processes a parametric function with the current input to the model, and the previous network state.
When the network is used for solving problems in a learning context, there is added an external aggregation function with the form:
\begin{equation}
    \label{OutputLayer}
    \y(t)=g(\x(t),\bm{\theta}_2), 
\end{equation}
where $g(\cdot)$ is also a parametric function where the parameters are collected in $\theta_2$.
The collection $\theta_1$ contains the weight matrix with input to the hidden neurons, and the weights of the hidden-to-hidden connections. The collection $\theta_2$ has the matrix with weights from hidden-to-hidden neurons to the output neurons~\cite{Martens2012}.
%
%
%The importance of network recurrences was analyzed in the seminal works of Elman, Hopfield and Jordan~\cite{Elman90}. In particular, the Elman work's is really nice and explicitly describes the difference between modeling a time-series with RNNs versus modeling using sliding windows and feedforwards.
Obviously, there are many other variations of RNN architectures, for more information we suggest~\cite{BengioDeepRNN}.

\section{Echo State Networks}
\subsection{Model specification}

\begin{figure}[h]
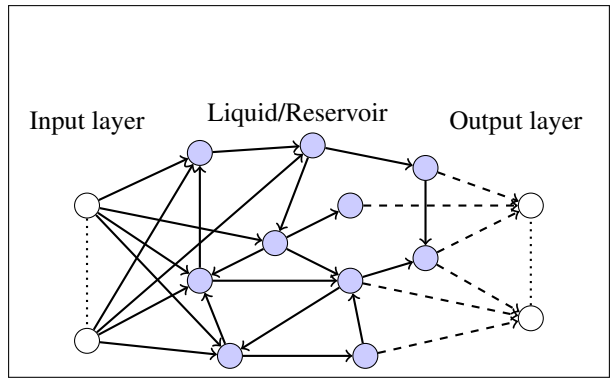

\begin{center}
\fbox{{\tikz[every node/.style={draw,circle}]{
{\node[fill=white]  (a) at (-1.5,3.1)[fill=white,draw=white,text=black] {Input layer};} 
{\node[fill=white]  (a) at (1.3, 3.2)[fill=white,draw=white,text=black] {Liquid/Reservoir};}
{\node[fill=white]  (a) at (4.2,3.1)[fill=white,draw=white,text=black] {Output layer};} 
{\node[fill=blue!20] (b) at (0.4,0) {};}
{\node[fill=blue!20] (c) at (2.2,0) {};}
{\node[fill=blue!20] (d) at (0,1) {};}
{\node[fill=blue!20] (e) at (1,1.5) {};}
{\node[fill=blue!20] (f) at (2,1) {};}
{\node[fill=blue!20] (g) at (0,2.7) {};}
{\node[fill=blue!20] (h) at (1.5,2.8) {};}
{\node[fill=blue!20] (i) at (2,2) {};}
{\node[fill=blue!20] (j) at (3,1.3) {};}
{\node[fill=blue!20] (k) at (3,2.5) {};}
{\node[fill=white] (o2) at (4.4,2) {};}
{\node[fill=white] (in1) at (-1.5,0.2) {};}
{\node[fill=white] (in2) at (-1.5,2) {};}
{\node[fill=white] (o1) at (4.4,0.5) {};}
{\draw (e)  edge[->,thick] (d) edge[->,thick] (i) edge[->,thick] (f);}
{\draw (d)  edge[->,thick] (g) edge[->,thick] (f);}
{\draw (f)  edge[->,thick] (j) edge[->,thick] (b);} 
{\draw (f) edge[->,thick,dashed] (o1);}
{\draw (b)  edge[->,thick] (c);}
{\draw (c)  edge[->,thick,dashed] (o1);}
{\draw (c)  edge[->,thick] (f);}
{\draw (g)  edge[->,thick] (h);}
{\draw (h)  edge[->,thick] (e) edge[->,thick] (k);}
{\draw (k) edge[->,thick] (j);}
{\draw (j)  edge[->,thick,dashed] (o1);}
{\draw (j)  edge[->,thick,dashed] (o2);}
{\draw (k)  edge[->,thick,dashed] (o2);}
{\draw (i)  edge[->,thick,dashed] (o2);}
{\draw (b)  edge[->,thick] (d);}
{\draw (in2)  edge[dotted,thick] (in1);}
{\draw (o1)  edge[dotted,thick] (o2);}
{\draw[->,black,thick] (in1)  edge[->,thick] (g);}
{\draw[->,black,thick] (in1)  edge[->,thick] (b);}
{\draw[->,black,thick] (in1)  edge[->,thick] (d);}
{\draw[->,black,thick] (in2)  edge [->,thick]  (g);}
{\draw[->,black,thick] (in2)  edge [->,thick]  (d);}
{\draw[->,black,thick] (in2)  edge [->,thick]  (e);}
{\draw[->,black,thick] (in1)  edge [->,thick] (h);}
{\draw[->,black,thick] (in2)  edge [->,thick] (b);}
}
}}
\caption{\label{FigESN} An example of the design of an Echo State Networks. Colored nodes form circuits and represent the reservoir nodes. The dashed edges are the only one with trainable weights.}
\end{center}
\end{figure}

The standard ESN model is a specific RNN case, then it has also the recurrent form~(\ref{recurrentState}), with the particular composition:
\begin{equation}
\label{ESNhiddenState}
\x(t)=f\big(\wi\va(t)+\wre\x(t-1)\big),
%\x(t)=f\big(\wi\va(t)+\wre\x(t-1)\big),
\end{equation}
where~$t$ is the discrete time index, $\wi$ is a $\Nx\times\Na$ matrix with input-to-hidden weights, the $\Nx\times\Nx$ matrix $\wre$ has the hidden-to-hidden weights. For simplicity, bias terms are omitted on the equations (common practice because it is straightforward to implement them by adding dummy terms).
The model output is given by:
\begin{equation}
\label{readout}
\vy(t)=g(\wo\x(t)),
\end{equation}
where $f(\cdot)$ is a contractive function (Lipschitz function) and $g(\cdot)$ is an activation function. In the case of the  $g(\cdot)$ as the identity function, then the readout expression~(\ref{readout}) is a linear model.
The recurrences have transitions given by~(\ref{ESNhiddenState}), at each time stamp the transition processes weighted information between the current state and the current input pattern. 
This transition function is fixed for all the learning dataset, due to the $\wi$ and $\wre$ are initialized and remain fixed during the training process. Therefore, expression~(\ref{ESNhiddenState}) is seen as a static random projection of the input space.
The projection sometimes is also referred as expansion because the dimension $\Nx$ is much larger than $\Na$ ($\Na\ll  \Nx$).
Only the parameters $\wo$ presented in~(\ref{readout}) are adjusted according to the learning problems.
A common way of solving readout equation~(\ref{readout}) is to apply a classical tool such as Tikhonov regularization~\cite{Jaeger09,Hart2020}.
Figures~\ref{ReservoirDiagrame} and~\ref{ReadoutDriagrame} illustrate the principle of ESN models. We can see the model as a pipeline with two consecutive independent blocks presented in Fig.~\ref{ReservoirDiagrame}) and Fig.~\ref{ReadoutDriagrame}).
Another illustrative figure of the RC principle is shown in~\ref{FigESN}. 
The reservoir nodes are illustrated by colored circles. Note, that there are circuits among the black nodes. The most common design of the ESN is to have a fully connected input-reservoir matrix, and also a fully connected reservoir-output matrix. Fig.~\ref{FigESN} is an illustrative simplification, and doesn't have a fully-connected input-reservoir nodes, neither reservoir-output nodes.
\begin{figure}
    \centering
\fbox{
    \tikz{
	      \node[draw,thick,fill=blue!9, rectangle,minimum size=4ex]  (id) at (0,0) {Reservoir};
      		\node[fill=white] (i) at (-3,0.8) {Input};
		\node[fill=white] (h) at (0,0.8) {Expansion};
		\node[fill=white] (g) at (3,0.8) {Projection};
		\node[fill=white] (a) at (-2.5,0) {$\vu(t)\in\R^{\Na}$};
     		\node[fill=white] (b) at (3,0)  {$\x(t)\in\R^{\Nx}$};
      		\node[fill=white] (wp) at (0,-0.9) {$\wre$};
  		\node[fill=white] (xt) at (1.9,-0.9) {$\x(t-1)$};
		\draw[->,black] (a)  edge[->,thick] (id);
		\draw[->,black] (id)  edge[->,thick] (b);
		\draw[->,black] (wp)  edge[->,thick] (id);
		\path[->,black,every loop/.style={looseness=10}] (b)  edge[->,thick,in=-90,out=-60] (id); 
      	}
}	
    \caption{Recurrent structure (reservoir) formalized in expression~(\ref{ESNhiddenState}). The reservoir is driven by an external input signal. Reservoir has randomly connected fixed weights that don't change in the learning process.\label{ReservoirDiagrame}}
    
    ~\\
    \bigskip
    
\fbox{
\tikz{
	      \node[draw,thick,fill=blue!9, rectangle,minimum size=4.5ex]  (id) at (0,0.3) {Supervised learning};
      		\node[fill=white] (i) at (-3,1.1) {Projection};
					\node[fill=white] (h) at (0,1.1) {Training};
					\node[fill=white] (g) at (2.5,1.1) {Output};
					\node[fill=white,text=blue] (a) at (-4,0.3) {$\x(t)\in\R^{\Nx}$};
     		\node[fill=white] (b) at (4,0.3)  {${\y}(t)\in\R^{\Ny}$};
      		\node[fill=white] (wp) at (0,-0.6) {{$\wo$}};
		\draw[->,black] (a)  edge[->,thick] (id);
		\draw[->,black] (id)  edge[->,thick] (b);
		\draw[->,black] (wp)  edge[->,thick] (id);
}
}
\caption{Readout structure formalized in expression~\ref{OutputLayer}. The projected points by the reservoir are taken as input of a supervised learning, e.g. linear regression. The parameters $\wo$ are trained using the empirical data.\label{ReadoutDriagrame}}
\end{figure}

\subsection{ESN variations}
The original model has been suffered some slight variations, we highlight the following ones(in some cases it has been investigated models that combine at the same type several of the variations presented here).
):
\begin{enumerate}
    \item Feedback connections~\cite{JaegerScience04}: 
\begin{equation}
\label{hiddenStateFeedback}
\x(t)=f\big(\wi\va(t)+\wre\x(t-1)+\wfb \y(t)\big),
%\x(t)=f\big(\wi\va(t)+\wre\x(t-1)\big),
\end{equation}
where $\wfb$ is a weight $\Nx\times\Ny$ matrix with the feedback connections going from the output neurons to the hidden structure. The feedback also can be integrated with an arbitrary delay.
\item Additional noise~~\cite{Rodan11}:
\begin{equation}
\label{hiddenStateNoise}
\x(t)=f\big(\wi\va(t)+\wre\x(t-1)+\noise(t)\big),
\end{equation}
to incorporate noise in the training procedures is a common used regularization strategy in NNs.
\item Leaky integrator ESN: the reservoir state variation is smoothed using a leaky rate~\cite{Jaeger09}:
\begin{equation}
\label{LI-ESN}
\x(t)= (1-\gamma)\x(t-1)+ \gamma \phi\big( \wi \va(t) +\wre \x(t-1)\big),
\end{equation}
where $\gamma\in(0,1]$ is the control parameter referred as {leaking rate}. It control the speed of the dynamic transitions, small values implies less sensitivity to the input pattern~\cite{Jaeger09}. 
\item Using ideas from recursive Self--Organizing Maps (SOMs)~\cite{MantasThesis,BasterCord11}, the neurons have the form:
$\bm{z}(t)=(z_1(t),\ldots,z_{\Nx}(t))$ through
\begin{equation}
\label{LI-ESN-SOM}
z_i(t)=\exp{\big(-\alpha||\wi_i-\va(t)||^2 -\beta ||\wre_i-\x(t-1)||^2\big)},
\end{equation}
for all $i\in\Nx$, and then updates its $\x(t)$ vector using leaky integrators:
\begin{equation}
\label{stateReservoir}
\x(t)=(1-\gamma)\x(t-1)+\gamma \bm{z}(t),
\end{equation}
where $\alpha$ and $\beta$ are the parameters to weight  the euclidean distance between inputs and $\wi$ and the previous state of the reservoir and $\wre$, respectively.
\end{enumerate}

Another type of variation are produced in the readout computation. It is common to see the aggregation of the recurrent state (expression~\ref{OutputLayer}) in linear form as follows:
\begin{equation}
\label{outputState}
\y(t)=g(\wo[\va(t);\x(t)]),
\end{equation}
where $\wo$ is the $\Nx\times (\Ny+\Na)$ reservoir-to-output weight matrix and $[\cdot;\cdot]$ denotes vector concatenation.
In addition, there exists the ESN with intrinsic plasticity~\cite{Schrauwen07}, where there are parameters in the activation functions of the reservoir units.
Furthermore, an ESN with two types of hidden-hidden connections (emulating excitatory and inhibitory signals) was developed in~\cite{Baster12ESQN,BasterrechPEIS2017}.
%

%backpropagation--decorrelation~\cite{Steil04}, decoupled ESN~\cite{Xue07}, leaky integrator~\cite{Jaeger07}, Evolino~\cite{Schmidhuber07}, etc.
%

\subsection{Properties}
An important property of the reservoir is the \textit{Echo State Property} (ESP) presented in~\cite{Jaeger01}.
In essence, the ESP guarantees that under certain algebraic conditions the reservoir state given by expression~(\ref{ESNhiddenState}) becomes asymptotically independent of the initial conditions ($\x(0)$)~\cite{Jaeger09}.
In the case of LSMs this property is presented under the name of \textit{fading memory}.
%
%The reservoir state does not asymptotically depend on the weights in the reservoir but rather just on the history of the input patterns. 
%
%In~\cite{Jaeger09} the property is presented as follows: 
%
%\begin{quotation}
%\textit{The effect of a previous state $\x(t)$ and a previous input $\va(t)$ on a future state $\x(t+k)$  should vanish gradually with $k\rightarrow \infty$.}
%\end{quotation}
In practical applications, it is enough to scale the reservoir weights for almost always ensured the ESP reservoir weight. Thus, the reservoir weights are typically scaled as $\ds{\wre:= \frac{\alpha\wre}{\rho(\wre)}}$, where $\alpha$ is a scaling parameter between $0$ and $1$.
The ESN property can be obtained even if  $\rho(\wre)>1$ for non--zero input or output feedback~\cite{MantasThesis, BasterrechIJCNN2017}. 
Nevertheless, the ESN is violated if $\rho(\wre)>1$ using the $\tanh(\cdot)$ activation function in the reservoir neurons and for zero input (no input)~\cite{Jaeger09}. In this situation, the internal state of the reservoir presents oscillations or even chaotic dynamics~\cite{Jaeger09}. 
Also, the property may be lost even if $\rho(\wre)<1$ although it is hard to build a reservoir where this occurs~\cite{Jaeger09}.
%
%\msb{MUY vago, necesito estudiar bien el paper Jaeger01 y meterme con lo de Sist. Dinamicos.}
%
%
\section{Applications of Evolutionary Algorithms on the RC area}
\label{EA}
Evolutionary techniques are helpful for improving NN architecture, and they can be applied for different purposes such as feature selection, weight tuning, graph design, parameter adaptation, rule extraction, and so on~\cite{Yao1999}.
Optimization of network weights using an evolutionary search were developed in EVOLINO~\cite{Schmidhuber2007}). In recent years, the interest has been augmented due to the fruitful novel areas of Neural Architecture Search and Neuroevolution~\cite{Elsken2019,StanleyNature2019}. 
%
%Several evolutionary methods have been developed to optimize NNs' architectures~\cite{Angeline1994,Yao1993,Schmidhuber2007,Rawal2016,Stanley2002}.

\subsection{What are the global parameters to be optimized?}
The RC family has also taken advantage of the recent advances in  evolutionary computation.
EA techniques have been used for tuning the following global parameters which affect the ESN's performance.
\bit
\item \textbf{Reservoir size}. In many ESN applications, it has been shown that the reservoir size influences the ESN accuracy. As in all learning systems, there is a tradeoff to reach in the size of the reservoir. If it is too small, we don't exploit enough the benefits of separating the set of possible inputs inside a larger space, and we can have poor training processes. If it is too large, training is easy, but generalizing can have issues, the usual \textit{over--fitting} phenomenon.
%
%the choice of the reservoir size must be a compromise between the training error minimization and the generalization ability of the ESN (\textit{to avoid overfitting}). Often in practice, the reservoir size can be about ten times bigger than the input dimension~\cite{Jaeger01}.
\item \textbf{Spectral radius of the reservoir weight matrix}. The spectral radius influences the memory capability of the model; a spectral radius $\rho(\wre)$ close to~$1$ is appropriate for learning tasks that require long memory, and a value close to~$0$ is adequate for tasks requiring short memory~\cite{Jaeger01}.
The role of the spectral radius is more complex when the reservoir is built with spiking neurons in the LSM model~\cite{Verstraeten07,Paugam09}.
\item  \textbf{Sparsity of the reservoir weight matrix}. It is recommended to define a sparse matrix $\wre$ with only between $15\%$ and $20\%$  of possible connections non-zero~\cite{Jaeger01}. The sparsity enables fast reservoir updates. In general, the matrices of input weights and feedback weights (in the case that, they exist) are complete.
\item \textbf{Injected noise}. The use of a noise vector for regularization requires some type of control of this noise.
\item \textbf{Parameter of the activation function}. The non-linearity of the reservoir projection is given by the type of activation function on the reservoir neurons.
\item \textbf{Leaky rate}. As was mentioned above, it controls the impact of the inputs in the dynamics, a type of speed control. It is a  scalar in $[0,1)$. 
\item  \textbf{Feedback connections}. A model with feedback connections is much more computational powerful, but the instability increases. For this reason, it is very hard problem to tune the vector with feedback connections~\cite{Lukosevicius2007Feedback}.
\item \textbf{Topology}. In spite of the large amount of works investigating the topological structure of the reservoir, a useful topology for significantly improving the performance of the ESN is still unknown. The classical approach consists of producing randomly connected reservoirs. 
%Although, it is unlikely that random connections 
\eit
\subsection{Summary of the literature}
Particle Swarm Optimization (PSO) and Genetic Algorithms (GAs) have been used as metaheuristics for finding the global parameters in~\cite{Anderson12,Ferreira11,Ferreira2013,Ferreira10}.
GAs also were used for finding the parameters of an extended ESN in medical applications
%Global parameters of an extended ESN (DRESN) were optimized using GAs for solving medical signal prognosis~\cite{Zhong2017}.
%
PSO was applied as a pre-training algorithm has been investigated in~\cite{BasterrechNabic14}, where the authors applied PSO for adjusting a subset of the reservoir weights. The authors instead of searching the global parameters, the paper works directly over a subset of reservoir weights.
In~\cite{Chatzidimitriou2010,Matzner2017} was studied growing RC topologies using Neat and HyperNEAT.
%
%In~\cite{Matzner2017} the evolution of the reservoir topology was made considering a fitness function that combines the memory capacities of the model and its accuracy.
%
Deep architectures, sometimes referred hierarchical reservoirs, that a kind of cascade of reservoir clusters where they are connected among is a new promising area inside the RC family~\cite{GallicchioTree,Gallicchio2017}.
GAs has been used for finding hyperparameters of hierarchical reservoirs and for tuning the weight  connections among the reservoirs~\cite{Qianli2020,Dale2018}. 
A variation of GA named microbial GA was used in~\cite{Dale2018}. 
Recently, PSO was introduced as a hybrid approach of swarm optimization and local search in hierarchical ESNs~\cite{Long2020}.
EvoESN was introduced in~\cite{BasterIJCNN2022,BasterGECCO21} where GA is used to optimize the weights of compressed reservoirs.
With the aim of finding good reservoir topology, particular cyclic graphs were analyzed in~\cite{Rodan2012,Rodan11}.
In~\cite{Schrauwen07, Steil2007}, reservoir units with parametric activation functions were trained for maximizing the entropy using the Intrinsic Plasticity (IP) rule.
Other approaches based on Evolutionary Algorithms (EAs) for optimizing hyper-parameters of the reservoir structure, such as number of reservoir units, scaling parameters, connectivity properties (density and spectral radius of the reservoir weights matrix), have also been introduced in the literature~\cite{Ferreira11,Ferreira2013}.
EAs have also been used for optimizing interconnected reservoirs~\cite{Dale2018,Qianli2020}, where GAs are applied for searching hyper-parameters of hierarchical reservoirs.
Besides, Bayesian optimization for finding  hyper-parameters of particular reservoirs was also analyzed in~\cite{Maat2018,Ribeiro2020}.
Furthermore, a relatively new prominent research in the RC area consists of models with interconnected reservoirs.
A tree structure where the tree-nodes satisfies the reservoir characteristics was developed in~\cite{GallicchioTree}. 
The weight connections among the reservoirs are set in order of creating a global contraction mapping.
Furthermore, layered/hierarchical architectures composed of multi-layered reservoirs have also been studied in~\cite{Qianli2020,Gallicchio2019b,Gallicchio2019, Dale2018,Gallicchio2017,GALLICCHIO201787}.
%\subsection{Hybrid techniques}

\subsection{Discussion about new trends and open problems}
\label{Discussion}
We have commented several important parameters of an RC model, and give some examples where EAs were used to optimize them.
However, different reservoir matrices with the same hyper-parameters may produce substantially different results~\cite{Schrauwen07}. 
In other words, choosing good global parameters' values at the moment of designing the model doesn't seem to be sufficient for fully exploiting the potential of ESNs.
We identify the following research directions of interest.
\begin{itemize}
    \item \textbf{Random projection dynamics}. There is a general consensus that the ESN model operates closes to the optimal situation when the projection is close to the edge of chaos~\cite{Jaeger09,Verstraeten07,Legenstein07}. There are some works of EAs applied to the Lyapunov exponents optimization. However, it is still possible to provide insights on the specific area of Lyapunov exponents estimation using EAs (for the particular case of dynamics coming from ESNs).
    \item \textbf{Quality of the projections}. When the model is used for forecasting and learning, then the projection should preserve some topological properties from the input space. A preliminary work was presented using Sammon mapping~\cite{BasterrechESANN2018b}, but we believe there are still a lot of approaches to be investigated.
    \item \textbf{Multi-objective approach}. In most of the works in the RC community, researchers apply EAs over quadratic errors as a measure of the prediction accuracy. However, there are other metrics for evaluating the good characteristics of the model such as memory capacity, robustness, complexity and speed. It would be an interesting direction apply multi-objective for optimizing several different metrics.
    \item \textbf{Self-organized reservoirs}. Even though was studied many years ago in~\cite{MantasThesis,Luko10,BasterCord11}. The area related to self-organization has grown up in the last years due to robotic applications, therefore most probably new developments can be done in this area.
    \item \textbf{Lifelong learning paradigm}. ESN model has very good properties for being used in continual learning. Parameters of control for the catastrophic forgetting problem, and other optimization problems of the lifelong learning area using RC techniques can be studied using EAs.
\end{itemize}

\section{Conclusions and future work}
This paper presents a brief survey of optmization on the RC paradigm.
%,
We presented the RC models optimized by EAs. 
As far as we are concerned, this manuscript is the first overview about this specific sub-family of the RC domain. 
We provided a comprehensive global overview of the RC literature that can help to the reader for future  implementation and studies of the RC and EA families. 
We described in detail the main parameters to be optimized, and provided a summary of the most important works of EAs used for finding good RC architectures.
We also discussed some drawbacks, opportunities and open questions for the research community.
This survey is not exhaustive, then the first work to be done in the close future is to improve the literature review and to collect new works referring the edge between EAs and RC.
%

%\clearpage
\bibliographystyle{plain}
\bibliography{References}
%\printbibliography
%%% Uncomment this line and comment out the ``thebibliography'' section below to use the external .bib file (using bibtex) .





% \end{thebibliography}

\end{document}

%% file: macrosBaster.tex
\newif\ifnotes\notestrue
\def\boxnote#1#2{\ifnotes\fbox{\footnote{\ }}\ \footnotetext{ From #1:
#2}\fi}
   %%%
   %%% --- modificaciones texto
\def\fgr#1{\boxnote{Seba}{\color{blue}#1}}
%\def\fsb#1{\boxnote{Sebastian}{\color{red}#1}}
   %%% --- comentarios en pie de pagina
\newcommand{\msb}[1]{{\color{blue}#1}}

\newcommand{\ben}{\begin{enumerate}}
\newcommand{\een}{\end{enumerate}}
\newcommand{\predLoad}{\langle\bm{\load}\rangle}
\newcommand{\grad}{\bm{\triangledown} E}

\newcommand{\bc}{\begin{center}}
\newcommand{\ec}{\end{center}}

\newcommand{\bit}{\begin{itemize}}
\newcommand{\eit}{\end{itemize}}

\newcommand{\ds}{\displaystyle}
\newcommand{\beq}{\begin{equation}}
\newcommand{\eeq}{\end{equation}}

\newcommand{\uu}[2]{(U_{#1}^{#2})_{0,0}}
\newcommand{\spec}{\mbox{sp}}
\newcommand{\rr}{\sqrt{pq}}

\newcommand{\vak}{\va^{(k)}}
\newcommand{\vbk}{\vb^{(k)}}

\newcommand{\ppij}{p^+_{i,j}}
\newcommand{\pnij}{p^-_{i,j}}

\newcommand{\lpi}{\lambda^+_{i}}
\newcommand{\lnni}{\lambda^-_{i}}
\newcommand{\lp}{\lambda^+}
\newcommand{\lnn}{\lambda^-}
\newcommand{\load}{\varrho}
\newcommand{\loadi}{\varrho_i}

\newcommand{\Tpi}{T^+_i}
\newcommand{\Tni}{T^-_i}

\newcommand{\Tp}{T^+}
\newcommand{\Tn}{T^-}

\newcommand{\wij}{w_{i,j}}
\newcommand{\wpij}{w^+_{i,j}}
\newcommand{\wnij}{w^-_{i,j}}
\newcommand{\wpji}{w^+_{j,i}}
\newcommand{\wnji}{w^-_{j,i}}

% Notacion de Reservoir computing.
%\newcommand{\x}{\mathbf{x}}
%\newcommand{\y}{\mathbf{y}}
%\newcommand{\vu}{\mathbf{u}}
\newcommand{\w}{\mathbf{W}}

\newcommand{\Mi}{\mathbf{M}^{\rm{in}}}
\newcommand{\Mr}{\mathbf{M}^{\rm{r}}}

\newcommand{\wn}{\mathbf{W}^{\rm{n}}}
\newcommand{\wre}{\mathbf{W}^{\rm{h}}}
\newcommand{\wi}{\mathbf{W}^{\rm{in}}}
\newcommand{\wo}{\mathbf{W}^{\rm{out}}}
\newcommand{\wfb}{\mathbf{W}^{\rm{fb}}}
\newcommand{\yt}{\y_{\text{target}}}

\newcommand{\va}{\mathbf{u}}
\newcommand{\vb}{\mathbf{b}}
\newcommand{\vx}{\mathbf{x}}
\newcommand{\vy}{\mathbf{y}}
\newcommand{\vw}{\mathbf{W}}
\newcommand{\vat}{\mathbf{a}(t)}
\newcommand{\vbt}{\mathbf{b}(t)}
\newcommand{\x}{\mathbf{x}}
\newcommand{\y}{\mathbf{y}}
\newcommand{\vs}{\mathbf{s}}
\newcommand{\Na}{n}
\newcommand{\Nb}{o}
\newcommand{\Ns}{p}

\newcommand{\vg}{\boldsymbol\gamma}

\newcommand{\cdI}{{I}}
\newcommand{\cdO}{{O}}
\newcommand{\X}{\mathbf{X}}
\newcommand{\Y}{\mathbf{Y}}
\newcommand{\p}{\mathbf{\mathbf{p}}}
\newcommand{\Prob}{\mathds{P}}
\newcommand{\N}{\mathds{N}}
\newcommand{\R}{\mathds{R}}
\newcommand{\gradientLb}{\bigtriangledown L(\bm{\beta})}

\newcommand{\SB}[1]{\textbf{\sf #1}}

\newcommand{\MOD}[1]{\textbf{\sf #1}}
\newcommand{\fpNoParen}[2]{\displaystyle{\frac{\partial{#1}}{\partial{#2}}}}
\newcommand{\fp}[2]{\displaystyle{\frac{\partial}{\partial{#2}}\bigg({#1}\bigg)}}

% Para el uso de index
\newcommand{\indice}[1]{#1\index{\sc{#1}}}

\newcommand{\indicat}[1]{\upharpoonleft\!\mid_{(#1)}}